%% file: main.tex
\documentclass[11pt]{article}
\usepackage[letterpaper,margin=1.25in]{geometry}
\usepackage[T1]{fontenc}
\usepackage[utf8]{inputenc}
\usepackage{lmodern}
\AtBeginDocument{\DeclareFontShape{T1}{lmr}{bx}{sc}{<->ssub*lmr/m/sc}{}}
\usepackage{amsmath,amsfonts,amssymb}
\usepackage{microtype}
\usepackage[numbers,compress]{natbib}
\usepackage{url}
\usepackage{float}
\usepackage{hyperref}
\hypersetup{hidelinks,
  pdftitle={Augmentron: Scalable Multi-View Visual Augmentation for Robot Learning},
  pdfauthor={Tanay Tandon and Aseem Doriwala and Jade Choghari and Catherine Weaver and Pragna Mannam}}
\usepackage[capitalise,nameinlink]{cleveref}
\crefname{appendix}{Section}{Sections}
\Crefname{appendix}{Section}{Sections}
\usepackage[table]{xcolor}
\usepackage{tikz}
\usetikzlibrary{patterns}
\usepackage{etoolbox}
\AtBeginEnvironment{tikzpicture}{%
  \renewcommand{\scriptsize}{\fontsize{7}{8}\selectfont}%
  \renewcommand{\tiny}{\fontsize{5}{6}\selectfont}}
\usepackage{colortbl,booktabs,multirow,graphicx,enumitem}
\newcommand{\hdr}[2]{\shortstack[c]{#1\strut\\{\tiny $n=#2$}}}

\title{\bfseries Augmentron: Scalable Multi-View Visual Augmentation for Robot Learning}
\date{}

\author{%
  {\normalsize\textsc{Tanay Tandon}$^{1,*,\dagger}$ \quad \textsc{Aseem Doriwala}$^{2,*}$ \quad \textsc{Jade Choghari}$^{2}$} \\
  {\normalsize\textsc{Catherine Weaver}$^{3}$ \quad \textsc{Pragna Mannam}$^{4}$} \\
  {\small $^{1}$University of Pennsylvania \quad $^{2}$Scale AI} \\
  {\small $^{3}$University of California, Berkeley \quad $^{4}$Scale AI, Inc.} \\[4pt]
  {\footnotesize $^{*}$Equal contribution.} \\
  {\footnotesize $^{\dagger}$Work done during an internship at Scale AI.}
}

\begin{document}
\maketitle

\begin{abstract}

Robot policies trained on visually narrow demonstrations can become sensitive
to environmental features that are unrelated to the task. Existing generative
augmentation methods can introduce visual diversity, but processing every frame
and camera view independently is computationally expensive and may produce
inconsistent observations across cameras and timestamps. We introduce Augmentron, a geometry-aware
pipeline that generates reusable scene representations and renders them through
recorded camera poses. This produces temporally and multi-view consistent
environment and workspace variations while preserving the robot, task objects,
actions, and annotations. Across three physical manipulation tasks, replacing
half of the real world training dataset with Augmentron data improves model success across three tasks for both in out-of-distribution distractor objects, and out-of-distribution workspace variation evaluations. In a separate comparison against alternate augmentation techniques, Augmentron achieves the highest pooled success in all three conditions. The results show that introducing augmented data to a data training recipe can help policies generalize to dynamic real-world settings.

\end{abstract}

\begin{figure}[H]
  \centering
  \includegraphics[width=\linewidth]{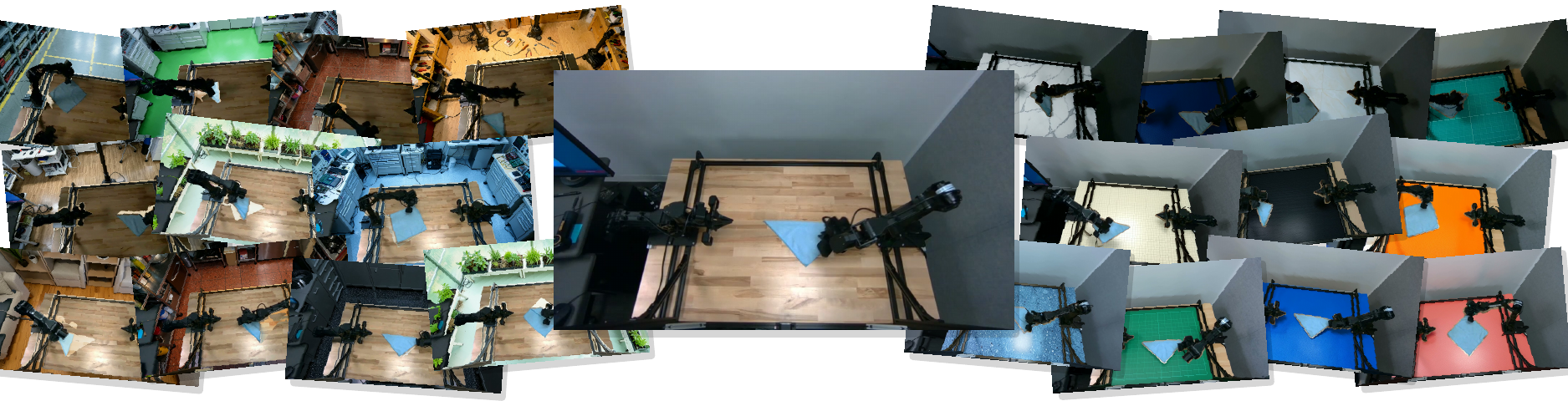}
  \caption{Environmental and workspace surface visual diversity produced by Augmentron: a geometry-aware pipeline that generates multi-view and temporally consistent data augmentation scalably for robotics manipulation datasets.}
  \label{fig:augmentation-diversity}
\end{figure}

\clearpage
\input{sections/01_intro}

\input{sections/02_related_work}
\input{sections/03_method}
\input{sections/04_experiments}
\input{sections/05_results_and_discussion}
\input{sections/06_conclusion}
\input{sections/07_appendix}

\end{document}

%% file: sections/01_intro.tex
\section{Introduction}

Robot demonstration datasets are often visually narrow: the same environments,
surfaces, and distractor configurations recur across many trajectories, even
though many of these visual features provide little information relevant to
the task. Policies trained on such data can fail when these details change
while the underlying task remains constant. In our evaluation, a real-only
policy fine-tuned from $\pi_{0.5}$~\citep{black2025pi05} achieves 66.7\% 
success on three in-distribution (ID) tasks, but success rate falls to 18.0\% with out-of-distribution (OOD) distractor objects. This
failure occurs despite broad vision-language-action pretraining
\citep{rt22023,kim2024openvla,octo2024}.

Collecting new demonstrations across diverse settings is expensive and
difficult to scale, making visual augmentation an attractive way to expand the
coverage of existing data. Augmenting robot trajectories, however, is more
difficult than editing individual images. The robot and task objects must be
preserved, and visual changes must remain consistent over time and across
camera views. Per-frame generative methods repeat segmentation and generation
for every observation, while video-generation costs grow with trajectory
length and camera count. Additionally, non-stationary robot wrist cameras introduce an additional
challenge due to their continuously changing viewpoints.

To address these challenges, we introduce \textsc{Augmentron}, a geometry-aware
pipeline that generates a reusable representation of a visual domain that avoids editing frames independently. \textsc{Augmentron} uses robot state and camera
calibration to initialize robot masks and amortizes segmentation across the
trajectory. Synthetically generated cubemaps and top-down scene images are then
reused across fixed and wrist camera views to create consistent environment
and working-surface variation. This design preserves the robot and task objects
while avoiding repeated per-frame generative inference.

We evaluate \textsc{Augmentron} across three physical manipulation tasks and
three visual conditions. In a controlled replacement experiment, substituting
half of the real world dataset with \textsc{Augmentron} data improves model performance across all tasks in
all three visual conditions. In a separate comparison of complete training volumes,
\textsc{Augmentron} achieves the highest overall task success rate in every condition,
including consistently stronger results than
RoboEngine~\citep{yuan2025roboengine}. \textsc{Augmentron} also
processes four-view demonstrations in $0.93\times$ their source-video duration,
compared with $16.28\times$ for RoboEngine on the same source set and hardware.

%% file: sections/02_related_work.tex
\section{Related Work}

\paragraph{Visual augmentation for robot learning.}
Prior work introduces visual diversity through chroma keying, semantic image
editing, and foreground-aware generation
\citep{teoh2024greenaug,chen2024semantic,yuan2025roboengine}. These methods show
that varying task-irrelevant appearance can improve policy robustness, but
typically process frames and camera views independently. \textsc{Augmentron} instead
reuses generated scene assets across an entire multi-view trajectory.

\paragraph{Video and multi-view generation.}
Video-diffusion methods condition generation on geometry, robot trajectories,
reference images, or simulation to produce temporally coherent demonstrations
\citep{liu2025robotransfer,dong2025emma,wang2026robovip,chen2026craft,nvidia2025cosmos}.
Their computational cost grows with trajectory length and camera count, while
consistency between generated views remains difficult. \textsc{Augmentron} takes a
rendering-oriented approach: calibration and robot state are used to project
reusable generated appearances into synchronized fixed and wrist-camera views.
We evaluate this tradeoff when fine-tuning a broadly pretrained policy
\citep{black2025pi05} under physical visual shifts.

%% file: sections/03_method.tex
\section{Method}
\label{sec:method}

\begin{figure}[t]
  \centering
  \includegraphics[width=\linewidth]{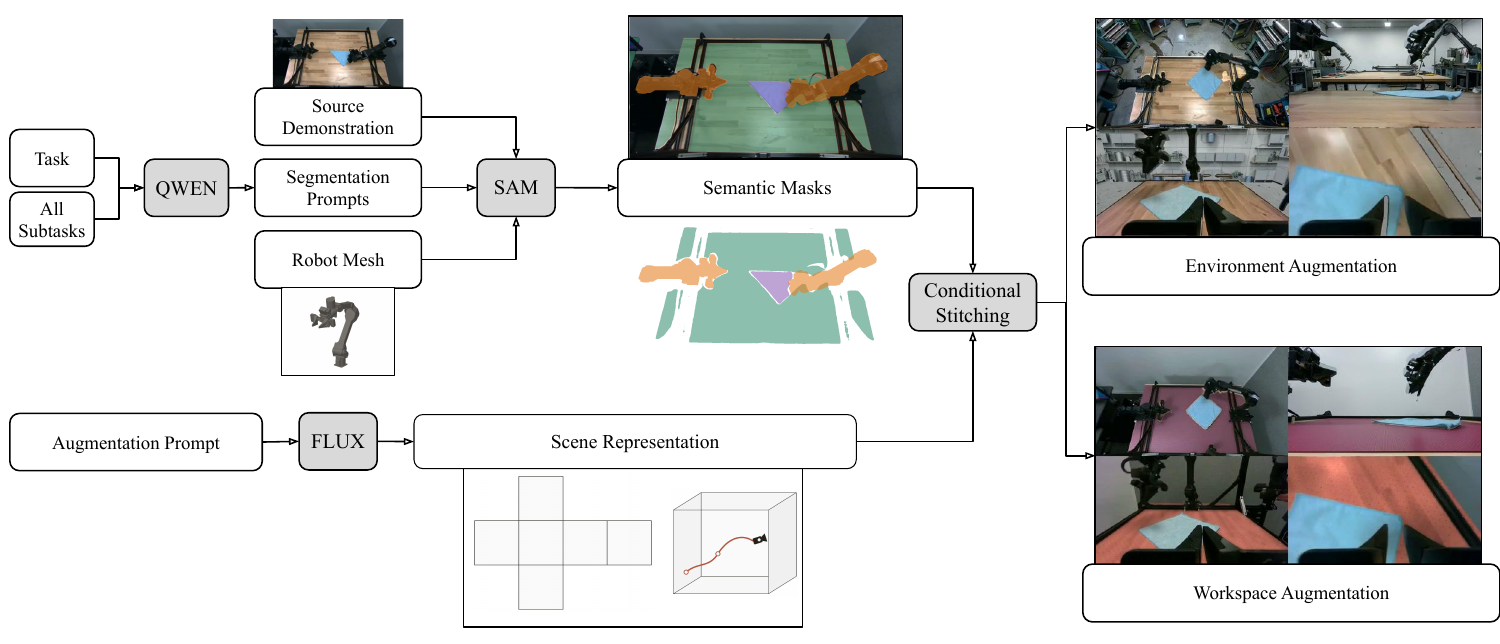}
  \caption{Overview of \textsc{Augmentron}. Task annotations and robot geometry
  identify regions that must remain unchanged. A generated scene representation
  is then rendered and composited with the source demonstration to produce
  environment or workspace augmentations.}
  \label{fig:pipeline-overview}
\end{figure}

\textsc{Augmentron} transforms the visual domain of a recorded demonstration
while preserving its robot state, actions, language labels, and timing.
As shown in \cref{fig:pipeline-overview}, the pipeline has three stages:
semantic mask generation, reusable scene representation generation, and
geometry-aware rendering.

\subsection{Semantic mask generation}

We first separate protected regions (robot and task-relevant objects) from parts of the scene that may be
modified during augmentation. The
workspace surface is segmented separately so that it can be preserved during
environment augmentation or replaced during workspace augmentation.

\paragraph{Robot masks.}
At each timestep, forward kinematics places the robot mesh in its recorded
configuration. The mesh is projected into each camera using the calibrated
camera parameters, producing a frame-specific robot mask. When necessary, this
geometric estimate is refined using segmentation.

\paragraph{Task objects and workspace.}
An instruction-tuned language model translates task and subtask annotations
into segmentation prompts, which are passed to SAM~3
\citep{carion2025sam3}. SAM~3 identifies both task-relevant objects and the
workspace surface. Fixed-camera surface masks can be reused over time because
their image-space position remains stable. Robot wrist-camera masks are estimated
separately due to their dynamic image-space position.

\subsection{Scene representation generation}

We construct an approximately foreground-free reference image by excluding
protected pixels from sampled frames and computing a pixelwise temporal median.
Because the robot and task objects move throughout the demonstration, locations
occluded in one frame are often visible in another. The median therefore
recovers much of the static scene while suppressing transient foreground
content.

The reference image and an augmentation prompt are passed to the FLUX diffusion
model family \citep{blackforestlabs2024flux}. FLUX produces a reusable scene
representation consisting of a static top-down scene image and a six-image
cubemap. The static image captures the working area, while the cubemap
represents the surrounding environment from each orthogonal direction. These
assets are reused across the full trajectory rather than regenerating each
observation independently.

\subsection{Geometry-aware rendering and stitching}

The generated scene representation is rendered using the recorded camera
geometry and combined with protected pixels from the original demonstration.
For a fixed camera, the static scene image is rendered once and reused across
timesteps. The original robot and task-object pixels are then composited over
the generated background using their semantic masks.

Wrist-camera views require rendering from a changing pose. Each camera pose is
recovered from the recorded robot state and calibrated transforms. For
workspace augmentation, the generated surface appearance is projected onto the
calibrated table plane and rendered from each wrist-camera viewpoint. For
environment augmentation, the cubemap is mapped onto the surrounding scene and
rendered from the same poses. Protected region pixels are then
composited over each rendering.

All synchronized camera streams use the same generated scene representation,
preserving temporal and multi-view consistency while leaving the original
actions and annotations unchanged.

%% file: sections/04_experiments.tex
\section{Experiments}
\label{sec:experiments}

Our experiments study whether training with \textsc{Augmentron} data improves physical
robustness over real-only training and how \textsc{Augmentron}
compares with generative and non-generative baselines.

\subsection{Tasks and policies}

We evaluate three manipulation tasks: placing a cup on a saucer, folding a towel,
and inserting a flower into a vase. The training set contains 1,793 real
demonstrations (9.16 hours) recorded with two fixed and two wrist cameras. The
policy observes the high fixed view and both wrist views. All policies are fine-tuned from $\pi_{0.5}$~\citep{black2025pi05} using the
LeRobot library~\citep{cadene2026lerobot}. The Real and \textsc{Augmentron} 50/50
policies share the same 30k-step training configuration, while the full
\textsc{Augmentron}, RoboEngine, and Masked Noise policies share the same 45k-step
configuration. Model initialization, batch size, camera inputs, and action
representation are otherwise held fixed. The demonstrations contain two fixed and two wrist-camera streams. \textsc{Augmentron} processes all four streams, while the policy observes only the high fixed view and both wrist views.

\subsection{Training comparisons}

Let $R$ denote the original real robot demonstration corpus, $E$ the \textsc{Augmentron}
environment-augmentation corpus, $T$ the \textsc{Augmentron} workspace-surface
augmentation corpus, $B$ the RoboEngine corpus, and $N$ the masked-noise
corpus. We define $1\times$ effective volume as one source-equivalent corpus,
corresponding to the complete set of real robot demonstrations and approximately 9.16
hours of total recorded data.

We evaluate five recipes: \textbf{Real} ($1R$; $1\times$),
\textbf{Augmentron 50/50} ($0.5R+0.25E+0.25T$; $1\times$),
\textbf{Augmentron 33/67} ($1R+1E+1T$; $3\times$),
\textbf{RoboEngine} ($1R+1B$; $2\times$)~\citep{yuan2025roboengine}, and
\textbf{Masked Noise} ($1R+1N$; $2\times$). In \textsc{Augmentron} 50/50, the real and
augmented halves use disjoint source episodes, so every source demonstration
is represented exactly once. The remaining augmented recipes retain the full
real corpus and add one or two source-equivalent corpora. Masked Noise assigns
one of five protected-region-preserving transformations shown in \cref{fig:noise-baseline} to each episode. Complete
recipe and corruption details appear in Appendix \cref{app:details}.

At 30k steps, we compare Real and \textsc{Augmentron} 50/50 for a controlled replacement
comparison because both effective data volume and training configuration are
matched. We additionally report the 30k checkpoint of \textsc{Augmentron} 33/67 as an
exploratory scaling condition. At 45k steps, we compare \textsc{Augmentron} 33/67, RoboEngine, and Masked Noise using the same optimization budget and training schedule. Effective data volume is not matched: \textsc{Augmentron} uses $3\times$ volume, while RoboEngine and Masked Noise use $2\times$. 

\begin{figure}[t]
  \centering
  \includegraphics[width=0.82\linewidth]{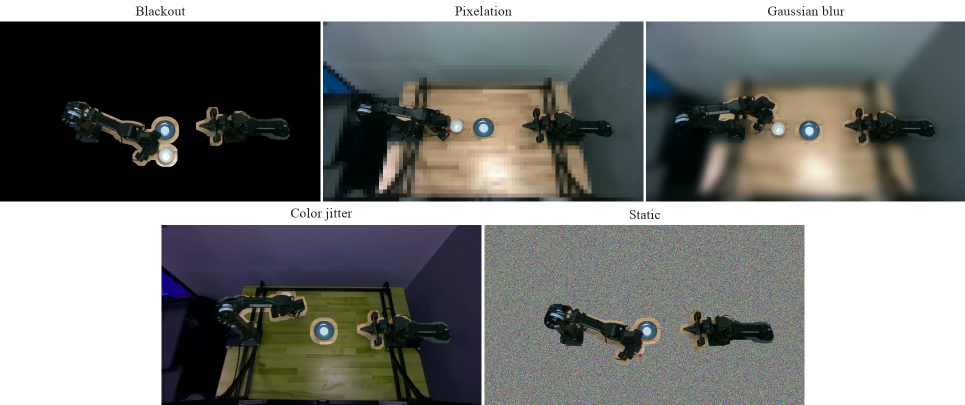}
  \caption{The five transformations used by the masked-noise baseline. Each
  corruption is applied outside the protected robot and task-object regions.}
  \label{fig:noise-baseline}
\end{figure}

\subsection{Physical evaluation}

Each policy is evaluated under three conditions: \emph{ID}, using the original
collection setup; \emph{OOD distractors}, adding task-irrelevant objects; and
\emph{OOD workspace}, replacing the visible wooden surface with a steel reflective surface. We run
50 physical rollouts for each complete task--policy--condition experiment. We report
task-level and pooled task success rates. Because there is one trained checkpoint
per task and recipe, rollout counts measure evaluation uncertainty but not
training-seed variation.

%% file: sections/05_results_and_discussion.tex
\section{Results and Discussion}
\label{sec:results}


\definecolor{augorange}{HTML}{ED6A0C}
\definecolor{chartgray}{HTML}{8C8C8C}

\newcommand{\resultaxes}[1]{%
  \foreach \y in {0,25,50,75,100} {%
    \draw[gray!18,line width=0.35pt] (0.45,\y) -- (#1,\y);%
    \node[anchor=east,text=gray!75,font=\scriptsize]
      at (0.35,\y) {\y};%
  }%
  \draw[gray!60,line width=0.45pt] (0.45,0) -- (#1,0);%
  \draw[gray!60,line width=0.45pt] (0.45,0) -- (0.45,100);%
}

\newcommand{\resultylabel}{%
  \node[rotate=90,font=\scriptsize]
    at (-0.58,50) {Success rate (\%)};%
}

\cref{tab:all-results} summarizes the task-level and pooled results. We first
analyze the controlled 30k-step replacement comparison, followed by the
exploratory scaling comparison and the 45k-step baseline comparison.

\begin{table}[t]
  \caption{Task and pooled success rates (\%). Bold marks the highest observed
  task success rate within each condition and training-step group. Each task cell
  contains 50 rollouts and each pooled cell contains 150.}
  \label{tab:all-results}
  \centering
  \scriptsize
  \setlength{\tabcolsep}{1.35pt}
  \renewcommand{\arraystretch}{1.04}

  \resizebox{\linewidth}{!}{%
  \begin{tabular}{
    rlcrrr>{\columncolor{gray!20}}r
    rrr>{\columncolor{gray!20}}r
    rrr>{\columncolor{gray!20}}r
  }
    \toprule
    & &
    & \multicolumn{4}{c}{ID}
    & \multicolumn{4}{c}{OOD distractors}
    & \multicolumn{4}{c}{OOD workspace} \\

    \cmidrule(lr){4-7}
    \cmidrule(lr){8-11}
    \cmidrule(lr){12-15}

    Steps
    & Recipe
    & \shortstack[c]{Effective\strut\\volume}
    & \hdr{Cup}{50} & \hdr{Towel}{50} & \hdr{Vase}{50} & \hdr{Pooled}{150}
    & \hdr{Cup}{50} & \hdr{Towel}{50} & \hdr{Vase}{50} & \hdr{Pooled}{150}
    & \hdr{Cup}{50} & \hdr{Towel}{50} & \hdr{Vase}{50} & \hdr{Pooled}{150} \\

    \midrule

    \multirow{3}{*}{30k}
      & Real
      & $1\times$
      & \textbf{86} & 68 & 46 & 66.7
      & 16 & 26 & 12 & 18.0
      & 78 & 54 & 20 & 50.7 \\

      & Aug. 50/50
      & $1\times$
      & 72 & \textbf{88} & \textbf{70} & \textbf{76.7}
      & 52 & 52 & \textbf{32} & 45.3
      & \textbf{80} & \textbf{94} & \textbf{28} & \textbf{67.3} \\

      & Aug. 33/67
      & $3\times$
      & 80 & 80 & 44 & 68.0
      & \textbf{76} & \textbf{64} & 12 & \textbf{50.7}
      & 62 & 88 & 26 & 58.7 \\

    \midrule

    \multirow{3}{*}{45k}
      & Aug. 33/67
      & $3\times$
      & 82 & 72 & \textbf{60} & \textbf{71.3}
      & \textbf{74} & 80 & \textbf{44} & \textbf{66.0}
      & 70 & \textbf{84} & \textbf{28} & \textbf{60.7} \\

      & RoboEngine
      & $2\times$
      & 70 & 82 & 20 & 57.3
      & 60 & 86 & 14 & 53.3
      & \textbf{78} & 40 & 6 & 41.3 \\

      & Masked Noise
      & $2\times$
      & \textbf{86} & \textbf{94} & 26 & 68.7
      & 62 & \textbf{88} & 6 & 52.0
      & 62 & 76 & 6 & 48.0 \\

    \bottomrule
  \end{tabular}%
  }
\end{table}

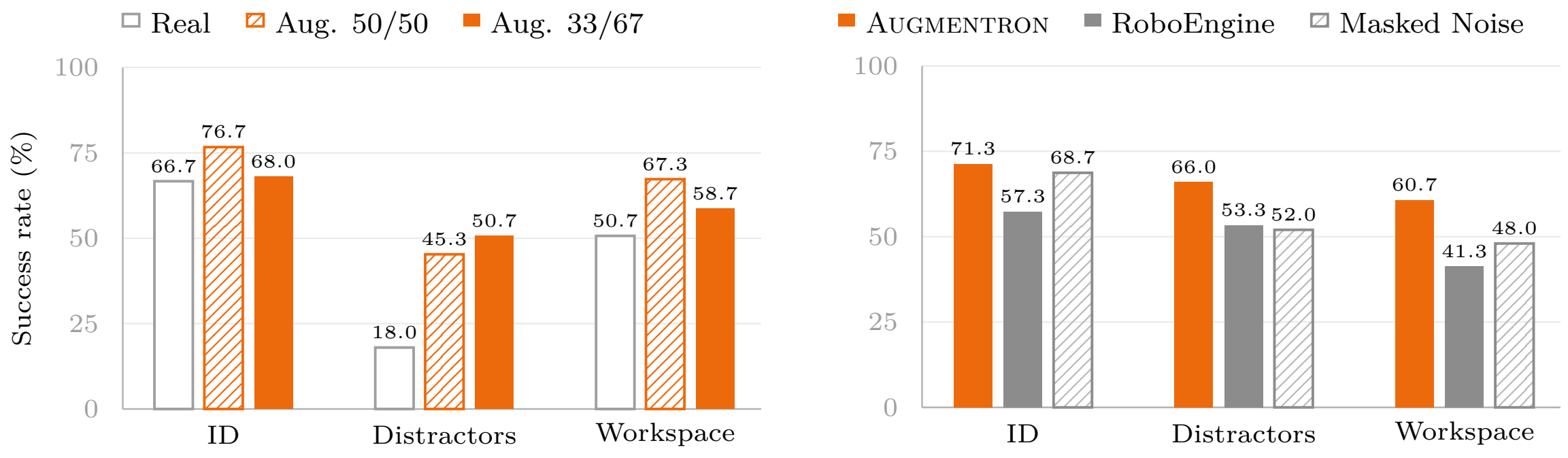
\begin{figure}[t]
  \centering

  \begin{minipage}[t]{0.49\linewidth}
    \centering

    {\scriptsize
      \tikz[baseline=-0.5ex]{
        \draw[gray!75,line width=0.7pt,fill=white]
          (0,0) rectangle (0.16,0.12);
      } Real
      \hspace{4pt}
      \tikz[baseline=-0.5ex]{
        \draw[
          augorange,
          line width=0.7pt,
          pattern=north east lines,
          pattern color=augorange
        ] (0,0) rectangle (0.16,0.12);
      } Aug. 50/50
      \hspace{4pt}
      \tikz[baseline=-0.5ex]{
        \fill[augorange] (0,0) rectangle (0.16,0.12);
      } Aug. 33/67
    }\par\vspace{1pt}

    \resizebox{\linewidth}{!}{%
    \begin{tikzpicture}[x=0.90cm,y=0.032cm,font=\scriptsize]
      \useasboundingbox (-0.85,-13) rectangle (7.15,106);
      \resultaxes{7.10}
      \resultylabel

      \draw[gray!75,line width=0.7pt,fill=white]
        (0.78,0) rectangle (1.18,66.7);
      \draw[
        augorange,
        line width=0.7pt,
        pattern=north east lines,
        pattern color=augorange
      ] (1.30,0) rectangle (1.70,76.7);
      \fill[augorange] (1.82,0) rectangle (2.22,68.0);

      \node[font=\tiny] at (0.98,71.2) {66.7};
      \node[font=\tiny] at (1.50,81.2) {76.7};
      \node[font=\tiny] at (2.02,72.5) {68.0};

      \draw[gray!75,line width=0.7pt,fill=white]
        (3.08,0) rectangle (3.48,18.0);
      \draw[
        augorange,
        line width=0.7pt,
        pattern=north east lines,
        pattern color=augorange
      ] (3.60,0) rectangle (4.00,45.3);
      \fill[augorange] (4.12,0) rectangle (4.52,50.7);

      \node[font=\tiny] at (3.28,22.5) {18.0};
      \node[font=\tiny] at (3.80,49.8) {45.3};
      \node[font=\tiny] at (4.32,55.2) {50.7};

      \draw[gray!75,line width=0.7pt,fill=white]
        (5.38,0) rectangle (5.78,50.7);
      \draw[
        augorange,
        line width=0.7pt,
        pattern=north east lines,
        pattern color=augorange
      ] (5.90,0) rectangle (6.30,67.3);
      \fill[augorange] (6.42,0) rectangle (6.82,58.7);

      \node[font=\tiny] at (5.58,55.2) {50.7};
      \node[font=\tiny] at (6.10,71.8) {67.3};
      \node[font=\tiny] at (6.62,63.2) {58.7};

      \node at (1.50,-7.5) {ID};
      \node at (3.80,-7.5) {Distractors};
      \node at (6.10,-7.5) {Workspace};
    \end{tikzpicture}%
    }

    {\scriptsize (a) 30k-step comparison}
  \end{minipage}
  \hfill
  \begin{minipage}[t]{0.49\linewidth}
    \centering

    {\scriptsize
      \tikz[baseline=-0.5ex]{
        \fill[augorange] (0,0) rectangle (0.16,0.12);
      } \textsc{Augmentron}
      \hspace{4pt}
      \tikz[baseline=-0.5ex]{
        \fill[chartgray] (0,0) rectangle (0.16,0.12);
      } RoboEngine
      \hspace{4pt}
      \tikz[baseline=-0.5ex]{
        \draw[
          chartgray,
          line width=0.7pt,
          pattern=north east lines,
          pattern color=gray!55
        ] (0,0) rectangle (0.16,0.12);
      } Masked Noise
    }\par\vspace{1pt}

    \resizebox{\linewidth}{!}{%
    \begin{tikzpicture}[x=0.90cm,y=0.032cm,font=\scriptsize]
      \useasboundingbox (-0.85,-13) rectangle (7.15,106);
      \resultaxes{7.10}

      \fill[augorange] (0.78,0) rectangle (1.18,71.3);
      \fill[chartgray] (1.30,0) rectangle (1.70,57.3);
      \draw[
        chartgray,
        line width=0.7pt,
        pattern=north east lines,
        pattern color=gray!55
      ] (1.82,0) rectangle (2.22,68.7);

      \node[font=\tiny] at (0.98,75.8) {71.3};
      \node[font=\tiny] at (1.50,61.8) {57.3};
      \node[font=\tiny] at (2.02,73.2) {68.7};

      \fill[augorange] (3.08,0) rectangle (3.48,66.0);
      \fill[chartgray] (3.60,0) rectangle (4.00,53.3);
      \draw[
        chartgray,
        line width=0.7pt,
        pattern=north east lines,
        pattern color=gray!55
      ] (4.12,0) rectangle (4.52,52.0);

      \node[font=\tiny] at (3.28,70.5) {66.0};
      \node[font=\tiny] at (3.80,57.8) {53.3};
      \node[font=\tiny] at (4.32,56.5) {52.0};

      \fill[augorange] (5.38,0) rectangle (5.78,60.7);
      \fill[chartgray] (5.90,0) rectangle (6.30,41.3);
      \draw[
        chartgray,
        line width=0.7pt,
        pattern=north east lines,
        pattern color=gray!55
      ] (6.42,0) rectangle (6.82,48.0);

      \node[font=\tiny] at (5.58,65.2) {60.7};
      \node[font=\tiny] at (6.10,45.8) {41.3};
      \node[font=\tiny] at (6.62,52.5) {48.0};

      \node at (1.50,-7.5) {ID};
      \node at (3.80,-7.5) {Distractors};
      \node at (6.10,-7.5) {Workspace};
    \end{tikzpicture}%
    }

    {\scriptsize (b) 45k-step comparison}
  \end{minipage}

  \caption{Pooled physical-rollout success. Left: 30k results; only Real
  versus \textsc{Augmentron} 50/50 is matched in data volume and training schedule.
  Right: 45k recipe comparison; \textsc{Augmentron} uses $3\times$ effective volume
  and each baseline uses $2\times$. Each bar aggregates
  \textbf{$n=150$} rollouts.}
  \label{fig:results-summary}
\end{figure}

\subsection{Does Augmentron improve over real-only training?}

The controlled comparison shows that replacing half of the real corpus with
\textsc{Augmentron} data improves performance across all three tested visual conditions
(\cref{fig:results-summary,tab:all-results}). Pooled success increases from
66.7\% to 76.7\% in ID, from 18.0\% to 45.3\% under OOD distractors, and from
50.7\% to 67.3\% under the workspace shift.

The largest improvement occurs under distractors. \textsc{Augmentron} 50/50 improves
cup-and-saucer success by 36 percentage points, towel folding by 26 points, and
vase insertion by 20 points. This pattern is consistent with augmentation
reducing sensitivity to visual features that repeatedly appear in the training
data but are not required to complete the task.

\subsection{Does augmented data ratio matter?}

The 30k scaling comparison does not show a uniform benefit from the larger
\textsc{Augmentron} 33/67 recipe. Relative to \textsc{Augmentron} 50/50, pooled success increases
from 45.3\% to 50.7\% under OOD distractors, but decreases from 76.7\% to
68.0\% in ID and from 67.3\% to 58.7\% under the workspace shift.

The task-level results are also mixed. Under distractors, \textsc{Augmentron} 33/67
improves cup-on-saucer and towel-folding performance, while vase insertion
falls from 32\% to 12\%. This comparison is exploratory: \textsc{Augmentron} 33/67 has
a larger effective data volume, and its 30k checkpoint comes from a run
scheduled for 45k steps. The observed differences therefore cannot be
attributed solely to the augmented-data fraction. Additional matched mixtures
and training seeds would be required to establish a scaling trend.

\subsection{How does \textsc{Augmentron} compare with the baselines?}

At 45k steps, \textsc{Augmentron} outperforms RoboEngine in the pooled comparison across
all three conditions (\cref{fig:results-summary}). It achieves 71.3\% in ID,
66.0\% under OOD distractors, and 60.7\% under the workspace variation, compared
with 57.3\%, 53.3\%, and 41.3\% for RoboEngine.

Masked Noise is a surprisingly competitive baseline, achieving 68.7\%,
52.0\%, and 48.0\% across the same conditions. RoboEngine and Masked Noise
produce similar pooled results under distribution shift, differing by 1.3
points under distractors and 6.7 points under the workspace shift. Both also
show substantially smaller ID to OOD-distractor drops than the 30k Real policy:
4.0 points for RoboEngine and 16.7 points for Masked Noise, compared with 48.7
points for Real. Because this comparison crosses training configurations, it
is descriptive rather than controlled, but it suggests that introducing visual
variation can reduce sensitivity to distractors.

Among the complete 45k recipes, \textsc{Augmentron} achieves the strongest pooled
results in every condition. These findings are consistent with a compound
benefit from its segmentation, richer scene representations, and temporal and
multi-view consistency. However, these components are evaluated jointly, and
\textsc{Augmentron} uses $3\times$ effective data volume compared with $2\times$ for
the baselines. The results therefore support the complete \textsc{Augmentron} recipe
without isolating the effect of any individual component.

\subsection{Runtime}

We measure end-to-end augmentation time on the same 30-minute, 104-episode
source set using one eight-H100 node. \textsc{\textsc{Augmentron}} processes four camera
views in 0.47 hours, corresponding to $0.93\times$ the duration of the source
videos. RoboEngine requires 8.14 hours, or $16.28\times$ source duration. A
measured one-view Cosmos-Transfer1 run~\citep{nvidia2025cosmos} requires
10.31 hours ($20.62\times$). Extrapolating its per-view stages to four views
gives 23.79 hours ($47.57\times$).

\textsc{Augmentron} avoids repeated generative inference for every frame. Once an
appearance asset has been generated, longer trajectories primarily add
rendering and compositing costs. This cost structure is particularly useful
for long demonstrations containing several synchronized camera streams.

%% file: sections/06_conclusion.tex
\section{Conclusion}

We presented \textsc{Augmentron}, a geometry-aware pipeline for producing
temporally and multi-view consistent visual augmentations of robot
demonstrations. Instead of invoking a generative model independently for every
frame, \textsc{Augmentron} reuses edited scene representations and renders them
through the recorded camera poses. This allows four-camera demonstrations to
be processed in near real time ($0.93\times$ the source video duration).

Replacing part of the real training corpus with \textsc{Augmentron} data
improved pooled performance across all three evaluation conditions relative to
real-only training. The pipeline also generated multi-view augmentations
substantially faster than the per-frame generative baselines evaluated in this
work. Although comparisons with other augmentation methods remain
task-dependent, these results show that reusable, geometry-grounded scene
assets can provide useful visual diversity at low processing cost.

Future work should evaluate additional tasks and visual shifts, train policies
across multiple seeds, and separately control for dataset size and optimization
exposure. Our results support \textsc{Augmentron} as an efficient and effective
augmentation approach in the settings studied, but do not establish universal
superiority over alternative methods.

\bibliographystyle{plainnat}
\bibliography{references}

\clearpage
\appendix

%% file: sections/07_appendix.tex
\section{Implementation and Evaluation Details}
\label{app:details}

\subsection{Mask generation}

\paragraph{Robot masks.}
Robot masks are generated by placing the calibrated robot mesh at each recorded
joint configuration and projecting it into each camera view. The experiments
use these projected masks directly, without learned refinement or additional
dilation.

\paragraph{Task-object masks.}
\texttt{Qwen/Qwen3-4B-Instruct-2507} converts task and subgoal annotations into
multiple SAM~3 prompts~\citep{carion2025sam3}. For fixed cameras, four sampled
frames are tiled into a temporal mosaic and segmented together. Wrist-camera
object masks are estimated separately for each view and frame because the
camera and task objects can move substantially in image space. Task-object
masks are dilated by six pixels before compositing to reduce the risk of
modifying object boundaries.

\paragraph{Workspace masks.}
Fixed-camera workspace masks are estimated at anchor frames and aggregated over
temporal windows. Wrist-camera workspace masks are generated from four-frame
temporal mosaics because the surface position changes in image space as the arm
moves.

\subsection{Appearance generation and rendering}

For each fixed camera, Augmentron constructs a reference image from up to 96
evenly spaced trajectory frames. Pixels covered by robot or task-object masks
are excluded, and a per-pixel temporal median is computed from the remaining
observations. Because the robot and objects move during a demonstration,
regions occluded in one frame are often visible in another. Regions that remain
unobserved are completed by the image-editing model.

We use \texttt{black-forest-labs/FLUX.2-klein-4B}
\citep{blackforestlabs2024flux} with four inference steps, guidance scale 1.0,
and a maximum image dimension of 1024 pixels. The resulting scene
representation consists of an edited static scene image and a six-face room
representation containing four walls, a floor, and a ceiling.

For environment augmentation, the room representation is rendered through each
recorded wrist-camera pose. For workspace augmentation, a sampled texture is
anchored to the calibrated workspace plane and projected into each view. The
original robot and task-object pixels are then composited over the rendered
scene. The same generated assets are reused throughout a trajectory, preserving
temporal and multi-view consistency.

Augmentron processes two fixed and two wrist-camera views. The policy consumes
the high fixed view and both wrist views; the low fixed view is augmented but
is not used as a policy input.

\subsection{Training recipe construction}

Let $R$ denote the real corpus, $E$ the Augmentron environment-augmentation
corpus, $T$ the Augmentron workspace-augmentation corpus, $B$ the RoboEngine
corpus, and $N$ the masked-noise corpus. One source-equivalent corpus is denoted
by $1\times$ effective volume.

\begin{itemize}[leftmargin=*,nosep]
    \item \textbf{Real} ($1R$; $1\times$): the complete real corpus.

    \item \textbf{Augmentron 50/50}
    ($0.5R+0.25E+0.25T$; $1\times$): the source episodes are divided into two
    disjoint halves. One half retains its real observations, while the other
    supplies the trajectories used for environment and workspace augmentation.
    Every source episode is therefore represented exactly once.

    \item \textbf{Augmentron 33/67}
    ($1R+1E+1T$; $3\times$): the complete real corpus plus one
    source-equivalent environment corpus and one source-equivalent workspace
    corpus.

    \item \textbf{RoboEngine}
    ($1R+1B$; $2\times$): the complete real corpus plus one source-equivalent
    corpus generated with RoboEngine~\citep{yuan2025roboengine}.

    \item \textbf{Masked Noise}
    ($1R+1N$; $2\times$): the complete real corpus plus one corrupted copy of
    every source trajectory.
\end{itemize}

All recipes retain coverage of every original source demonstration. They differ
in visual composition and effective data volume.

\subsection{Dataset and policy training}

The real corpus contains 1,793 demonstrations totaling approximately
9.16 hours: 527 cup-and-saucer demonstrations, 868 towel-folding
demonstrations, and 398 vase-insertion demonstrations. The complete corpus is
stored in the \texttt{LeRobotDataset} format~\citep{cadene2026lerobot}. Images
are recorded at $1280\times720$ and resized to $224\times224$ for training.

All policies are fine-tuned from the same $\pi_{0.5}$
initialization~\citep{black2025pi05}. The policy receives three camera views
and a 14-dimensional robot state and predicts 14-dimensional actions. Actions
are represented relative to the current state, except for the gripper
dimension.

We train the complete model, including the vision encoder, using AdamW with a
learning rate of $2.5\times10^{-5}$, $\beta_1=0.9$, $\beta_2=0.95$,
$\epsilon=10^{-8}$, weight decay $0.01$, and gradient clipping at 1.0. The
learning rate follows a cosine schedule with 1,000 warmup steps and a minimum
of $2.5\times10^{-6}$. Training uses a batch size of 32 on each of eight ranks,
giving a global batch size of 256. All runs use seed 1000.

The controlled Real and Augmentron 50/50 runs use a 30k-step cosine-decay
horizon. The reported 30k Augmentron 33/67 result is an intermediate checkpoint
from a run scheduled for 45k steps and therefore follows a 45k-step decay
schedule. All policies in the 45k baseline comparison use the 45k-step
schedule. We train one checkpoint for each task and recipe.

\subsection{Masked-noise baseline}

Masked Noise applies one of five transformations outside the protected robot
and task-object regions:

\begin{itemize}[leftmargin=*,nosep]
    \item random RGB static;
    \item pixelation using $16$-pixel blocks;
    \item Gaussian blur with $\sigma=18$;
    \item complete blackout; and
    \item color jitter.
\end{itemize}

A corruption family is assigned deterministically to each episode and retained
across all frames and policy views. Assignments are balanced by frame count
within each task. Random parameters are generated deterministically from the
episode, camera view, and frame identifiers.

\subsection{Physical evaluation protocol}

Each policy is evaluated under three conditions: \emph{ID}, using the original
collection setup; \emph{OOD distractors}, with task-irrelevant objects added to
the scene; and \emph{OOD workspace}, with the visible wooden working surface
replaced by steel.

We conduct 50 physical rollouts for each complete
task--policy--condition cell, yielding 150 rollouts for each pooled result.
Starting configurations are manually reset but are not paired across policies.
Each rollout lasts at most 1,200 control steps, corresponding to 40 seconds at
30~Hz. The operator assigns a score from 0 to 3 to denote progress, and a score of 3 means task success. Trials interrupted by robot or operator faults are voided
and repeated.

\subsection{Runtime benchmark protocol}

Runtime is measured on the same 30-minute cup-and-saucer source subset,
containing 104 episodes, 54,000 frames at 30~Hz, and four camera views. Each
method is run on one \texttt{a3.megagpu.8g} node with eight NVIDIA H100 GPUs.
Timing begins with cold model construction and ends after the local output
audit. It excludes queueing, container-image retrieval, credential and GPU
gates, artifact staging, inter-boot delays, and result publication.

Augmentron processes the source data in 0.47 hours, or $0.93\times$ its
source-video duration. RoboEngine requires 8.14 hours, or $16.28\times$ source
duration. A measured single-view Cosmos-Transfer1 run requires 10.31 hours
($20.62\times$). Extrapolating its per-view stages gives an estimated
four-view runtime of 23.79 hours ($47.57\times$); this four-view value was not
measured directly.